\documentclass[runningheads]{llncs}

\usepackage{eccv}

\usepackage{eccvabbrv}

\usepackage{graphicx}
\usepackage{booktabs}

\usepackage{url}            % simple URL typesetting

\usepackage{amsfonts}       % blackboard math symbols
\usepackage{nicefrac}       % compact symbols for 1/2, etc.
\usepackage{xcolor}         % colors
\usepackage{enumitem}
\usepackage{amsmath}
\usepackage{wrapfig}        
\usepackage{gradient-text}
\usepackage{multirow}
\usepackage{algorithm}    
\usepackage{algpseudocode}
\usepackage{colortbl}
\usepackage{soul}
\usepackage{caption}
\usepackage{makecell}
\usepackage{pifont}
\usepackage[normalem]{ulem}

\usepackage[accsupp]{axessibility}  % Improves PDF readability for those with disabilities.

\usepackage{hyperref}

\usepackage{orcidlink}

\begin{document}

% ---------------------------------------------------------------
% TODO REVIEW: Replace with your title
\title{ExploreVLA: Dense World Modeling and Exploration for End-to-End Autonomous Driving} 

% TODO REVIEW: If the paper title is too long for the running head, you can set
% an abbreviated paper title here. If not, comment out.
\titlerunning{ExploreVLA}

% TODO FINAL: Replace with your author list. 
% Include the authors' OCRID for the camera-ready version, if at all possible.

\author{Zihao Sheng\inst{1,2}\orcidlink{0000-0002-1095-2422}\textsuperscript{*} 
\and
Xin Ye\inst{1} %\textsuperscript{$\dagger$} 
\and
Jingru Luo\inst{1} 
\and
Sikai Chen\inst{2}\orcidlink{0000-0002-5931-5619}%\textsuperscript{$\ddagger$} 
\and
Liu Ren\inst{1}
}

% TODO FINAL: Replace with an abbreviated list of authors.
\authorrunning{Z.~Sheng et al.}
% First names are abbreviated in the running head.
% If there are more than two authors, 'et al.' is used.

% TODO FINAL: Replace with your institution list.
\institute{Bosch Research North America \& Bosch Center for Artificial Intelligence (BCAI) \and
University of Wisconsin--Madison \\
\email{\{zihao.sheng,sikai.chen\}@wisc.edu} \\
\email{\{xin.ye3,jingru.luo,liu.ren\}@us.bosch.com} \\
}

\maketitle

{\renewcommand{\thefootnote}{}%
\footnotetext{\textsuperscript{*}Work was done during internship at Bosch, supervised by Xin Ye}
}

\begin{abstract}
End-to-end autonomous driving models based on Vision-Language-Action (VLA) architectures have shown promising results by learning driving policies through behavior cloning on expert demonstrations. However, imitation learning inherently limits the model to replicating observed behaviors without exploring diverse driving strategies, leaving it brittle in novel or out-of-distribution scenarios. Reinforcement learning (RL) offers a natural remedy by enabling policy exploration beyond the expert distribution. Yet VLA models, typically trained on offline datasets, lack directly observable state transitions, necessitating a learned world model to anticipate action consequences. In this work, we propose a unified understanding-and-generation framework that leverages world modeling to simultaneously enable meaningful exploration and provide dense supervision. Specifically, we augment trajectory prediction with future RGB and depth image generation as dense world modeling objectives, requiring the model to learn fine-grained visual and geometric representations that substantially enrich the planning backbone. Beyond serving as a supervisory signal, the world model further acts as a source of intrinsic reward for policy exploration: its image prediction uncertainty naturally measures a trajectory's novelty relative to the training distribution, where high uncertainty indicates out-of-distribution scenarios that, if safe, represent valuable learning opportunities. We incorporate this exploration signal into a safety-gated reward and optimize the policy via Group Relative Policy Optimization (GRPO). Experiments on the NAVSIM and nuScenes benchmarks demonstrate the effectiveness of our approach, achieving a state-of-the-art PDMS score of 93.7 and an EPDMS of 88.8 on NAVSIM.
The code is available at \url{https://zihaosheng.github.io/ExploreVLA/}.
\end{abstract}

\section{Introduction}
\label{sec:intro}

End-to-end autonomous driving has advanced rapidly with the emergence of Vision-Language-Action (VLA) architectures~\cite{hwang2024emma,jiang2025diffvla,zhou2025opendrivevla,wang2025alpamayo}, which unify perception, reasoning, and planning within a single model.
By leveraging the representational power of large vision-language models, these models have demonstrated promising capabilities in translating raw sensor observations into driving actions.
Yet, the dominant training paradigm for such models (behavior cloning on expert demonstrations or supervised fine-tuning) introduces a fundamental bottleneck: the learned policy can only replicate the behaviors it has observed, without the ability to discover alternative strategies that may be equally or more effective.
This limitation manifests as distributional brittleness: when confronted with scenarios that deviate from the expert distribution, the policy lacks the exploratory experience needed to generalize~\cite{ross2011reduction,codevilla2019exploring}.

\begin{figure}[tb]
  \centering
  \includegraphics[width=0.99\textwidth]{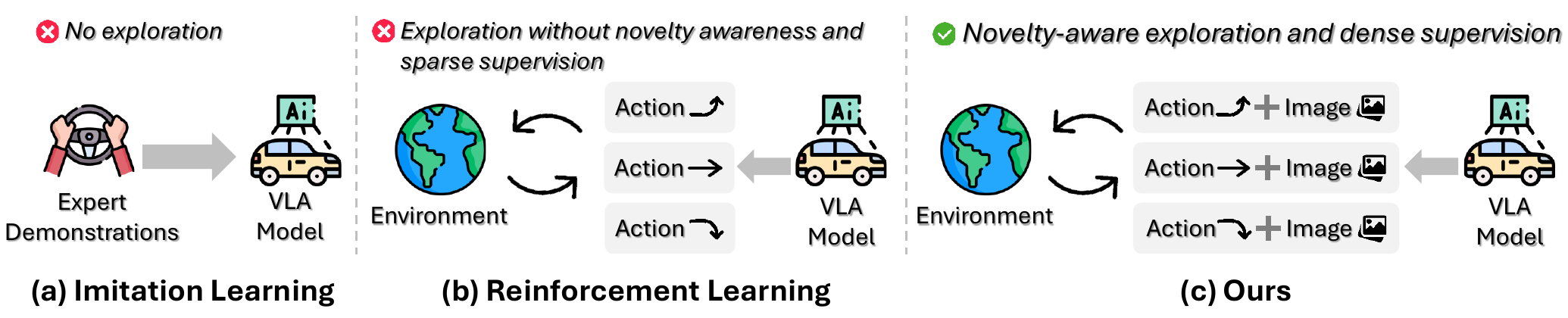}
  \caption{\textbf{Comparison of training paradigms for VLA-based autonomous driving.} (a) Imitation learning directly clones expert demonstrations without exploration. (b) Previous reinforcement learning enables policy exploration, but cannot distinguish expert imitation from genuine out-of-distribution discovery and relies on sparse supervision. (c) Our approach augments RL with dense world modeling supervision via future image generation, while leveraging image prediction uncertainty as a novelty measure to identify and prioritize valuable exploratory strategies.}
  \label{fig:paradigm}
\end{figure}

Reinforcement learning (RL) offers a principled mechanism to overcome this limitation by allowing the agent to explore beyond the boundaries of expert data and optimize its policy through trial-and-error interaction~\cite{guo2025improving}.
Recent advances in RL post-training, exemplified by Group Relative Policy Optimization (GRPO)~\cite{shao2024deepseekmath}, have demonstrated that sampling diverse candidate outputs and performing relative ranking can effectively improve policy quality atop strong pretrained models. However, applying RL to autonomous driving poses challenges distinct from other domains. In language tasks, state transitions are fully determined by the model's output, and outcomes are immediately observable. In robotics, high-fidelity simulators enable safe trial-and-error. Autonomous driving enjoys neither advantage: the consequences of a planned trajectory depend on complex scene dynamics that no existing simulator faithfully captures. These challenges motivate the need for a world model that can internalize environment dynamics and anticipate action consequences from data alone. Yet a further challenge remains: standard task-level rewards such as Predictive Driver Model Score (PDMS)~\cite{dauner2024navsim} evaluate trajectory quality but do not distinguish between policies that merely replicate expert behavior and those that have genuinely discovered novel strategies. An exploration signal orthogonal to task performance is therefore essential to unlock the full potential of RL post-training.

A second limitation of existing VLA driving models is the sparsity of their supervisory signals.
Most approaches rely on textual descriptions and trajectory waypoints as training targets~\cite{zhou2025autovla,zhou2025opendrivevla}, which, while informative for high-level decision-making, fail to capture the rich spatial geometry and fine-grained appearance of driving scenes.
This supervisory deficit constrains the model's ability to build comprehensive representations, particularly for aspects of the environment (such as road topology, object extent, and depth ordering) that are critical for safe planning but are not explicitly encoded in sparse action labels~\cite{li2025drivevla}.

In this work, we propose a unified understanding-and-generation framework that addresses both limitations through a single mechanism: \textbf{dense world modeling and exploration} (\cref{fig:paradigm}).
Specifically, we augment trajectory prediction with future RGB and depth image generation as auxiliary objectives.
On the supervision side, these generation tasks require the model to predict fine-grained visual appearance and metric geometry of future scenes, providing dense gradient signals that substantially enrich the planning backbone's visual and geometric representations.
On the exploration side, we leverage a key insight: \emph{the world model's image prediction uncertainty naturally measures a trajectory's novelty relative to the training distribution}.
Since the world model is trained exclusively on expert demonstrations, it produces low-uncertainty predictions for trajectories within the expert distribution but exhibits high uncertainty on out-of-distribution (OOD) trajectories. Crucially, this uncertainty reflects distance to the \emph{entire} training distribution, rather than deviation from a single ground-truth trajectory, which provides a more reliable novelty signal than trajectory distance alone.
We incorporate this signal into a safety-gated exploration reward: trajectories that achieve high PDMS scores while exhibiting high prediction uncertainty are identified as valuable discoveries of new successful strategies and receive an exploration bonus.
This composite reward is optimized via GRPO, enabling the policy to expand its behavioral repertoire while maintaining driving safety.

Our contributions can be summarized as follows:
\begin{itemize}
    \item We introduce a novel exploration mechanism for RL post-training that uses the world model's image prediction uncertainty as an intrinsic novelty measure, coupled with a safety-gated reward to encourage beneficial out-of-distribution exploration.
    \item We propose a unified VLA framework that jointly predicts future trajectories, RGB images, and depth images, leveraging dense world modeling to provide rich visual and geometric supervision for the planning backbone.
    \item We demonstrate state-of-the-art performance on the NAVSIM benchmark with a PDMS of 93.7 and an EPDMS of 88.8, and validate the generalizability of our approach on the nuScenes dataset.
\end{itemize}

\section{Related Work}

\subsection{World Models for Autonomous Driving}
In the context of autonomous driving, world models offer the ability to predict future states of the driving scene, enabling planning, data augmentation, and closed-loop evaluation without costly real-world interactions~\cite{guan2024world,feng2025survey,ding2025understanding,yan2025ad}.
A significant line of work focuses on video prediction-based world models~\cite{gao2024vista,hassan2025gem,yang2025resim}. For example, GAIA-1~\cite{hu2023gaia} leverages a generative model conditioned on video, text, and action inputs to produce realistic driving videos. DriveDreamer~\cite{wang2024drivedreamer} generates future driving frames conditioned on HDMaps and 3D bounding boxes. GenAD~\cite{yang2024generalized} proposes an action-conditioned video generation framework that supports long-horizon future prediction for planning. 
Another prominent direction explores world models within structured geometric spaces. UniWorld~\cite{min2023uniworld} proposes a unified framework for 4D occupancy forecasting, and OccWorld~\cite{zheng2024occworld} extends this idea by predicting 3D occupancy and ego-motion jointly, providing a compact world representation for downstream planning. UniScene~\cite{li2025uniscene} further scales this paradigm by jointly generating consistent future semantic occupancy, LiDAR and multi-view images. 
Beyond generation and prediction, several works integrate world models into the planning pipeline. For instance, WoTE~\cite{li2025end} incorporates a BEV world model to predict future states for online trajectory evaluation and selection. OmniNWM~\cite{li2025omninwm} employs a learned navigation world model to generate imagined futures and derive planning rewards from predicted scene dynamics. World4Drive~\cite{zheng2025world4drive} further introduces a latent world model that predicts future latent states under multiple driving intentions and selects trajectories via a learned selector. In our work, we leverage the world model not only for future state prediction but also as a source of intrinsic reward signals that encourage the policy to explore diverse and informative driving behaviors, thereby improving generalization beyond the coverage of the training data.

\subsection{VLA Models for Autonomous Driving}
The integration of vision, language, and action within a unified framework has emerged as a promising paradigm for autonomous driving~\cite{jiang2025survey,li2025recogdrive}. Early efforts, such as DriveGPT-4~\cite{xu2024drivegpt4}, use frozen VLMs to narrate driving scenes but do not directly output control signals. Subsequent modular VLA approaches began embedding language into the planning loop. For example, OpenDriveVLA~\cite{zhou2025opendrivevla} fuses multimodal sensor inputs with textual route instructions to generate interpretable waypoints, while RAG-Driver~\cite{yuan2024rag} introduces retrieval-augmented planning for long-tail scenarios. The field then advances toward unified end-to-end architectures. EMMA~\cite{hwang2024emma} jointly performs detection and planning within a single VLM, and DiffVLA~\cite{jiang2025diffvla} combines diffusion-based trajectory sampling with language-conditioned embeddings. Most recently, reasoning-augmented VLA models have pushed the frontier further. ORION~\cite{fu2025orion} incorporates a transformer memory module for long-horizon reasoning, and AutoVLA~\cite{zhou2025autovla} fuses CoT reasoning and trajectory planning in a single autoregressive transformer. Alpamayo-R1~\cite{wang2025alpamayo} introduces causally grounded Chain-of-Causation reasoning tightly integrated with trajectory prediction, enhancing reasoning-action consistency and long-tail safety performance.
Despite this rapid progress, most existing VLA models rely on textual descriptions and action trajectories as the primary supervisory signal, which are inherently sparse. This supervisory sparsity limits the model's ability to learn comprehensive scene representations. In contrast, our work leverages RGB and depth images as auxiliary dense supervisory signals to encourage the model to capture richer visual and geometric cues, thereby yielding more accurate and robust trajectory planning.

\subsection{Unified Understanding and Generation Models}
In foundation model research, the long-standing separation between understanding and generation has motivated a growing effort to unify both within a single architecture for greater scalability and cross-task synergy~\cite{xie2024show,chen2025janus,wang2026multimodal}. In autonomous driving, this paradigm has gained significant traction as researchers seek to bridge the gap between world modeling and end-to-end planning. For example, FutureSightDrive~\cite{zeng2025futuresightdrive} proposes a spatio-temporal visual Chain-of-Thought framework where a VLA model first generates future frames, including lane lines, 3D bounding boxes, and complete future images, as visual intermediate reasoning steps, then predicts trajectories conditioned on these imagined futures. Policy World Model (PWM)~\cite{zhao2025pwm} pre-trains on large-scale action-free video generation to learn world dynamics, then fine-tunes with a collaborative formulation where trajectory planning is explicitly conditioned on forecasted future states. DriveVLA-W0 ~\cite{li2025drivevla} introduces world modeling objectives to unlock data scaling laws for end-to-end driving. UniDrive-WM~\cite{xiong2026unidrive} unifies scene understanding, trajectory planning, and trajectory-conditioned future image generation within a single VLM. Epona~\cite{zhang2025epona} combines autoregressive causal modeling with diffusion-based generation through decoupled spatiotemporal factorization, enabling both high-fidelity video synthesis and trajectory planning. Together, these works demonstrate that jointly modeling future scene generation and action prediction yields more informed and anticipatory planning. Our work follows this unified paradigm by leveraging RGB and depth image generation as dense supervisory signals alongside trajectory prediction, while further employing the world model to provide intrinsic reward signals that encourage exploratory driving behaviors and improve generalization beyond the training distribution.

\section{Methodology}\label{sec:methodology}

\subsection{Overview}
\begin{figure}[tb]
  \centering
  \includegraphics[width=0.9\textwidth]{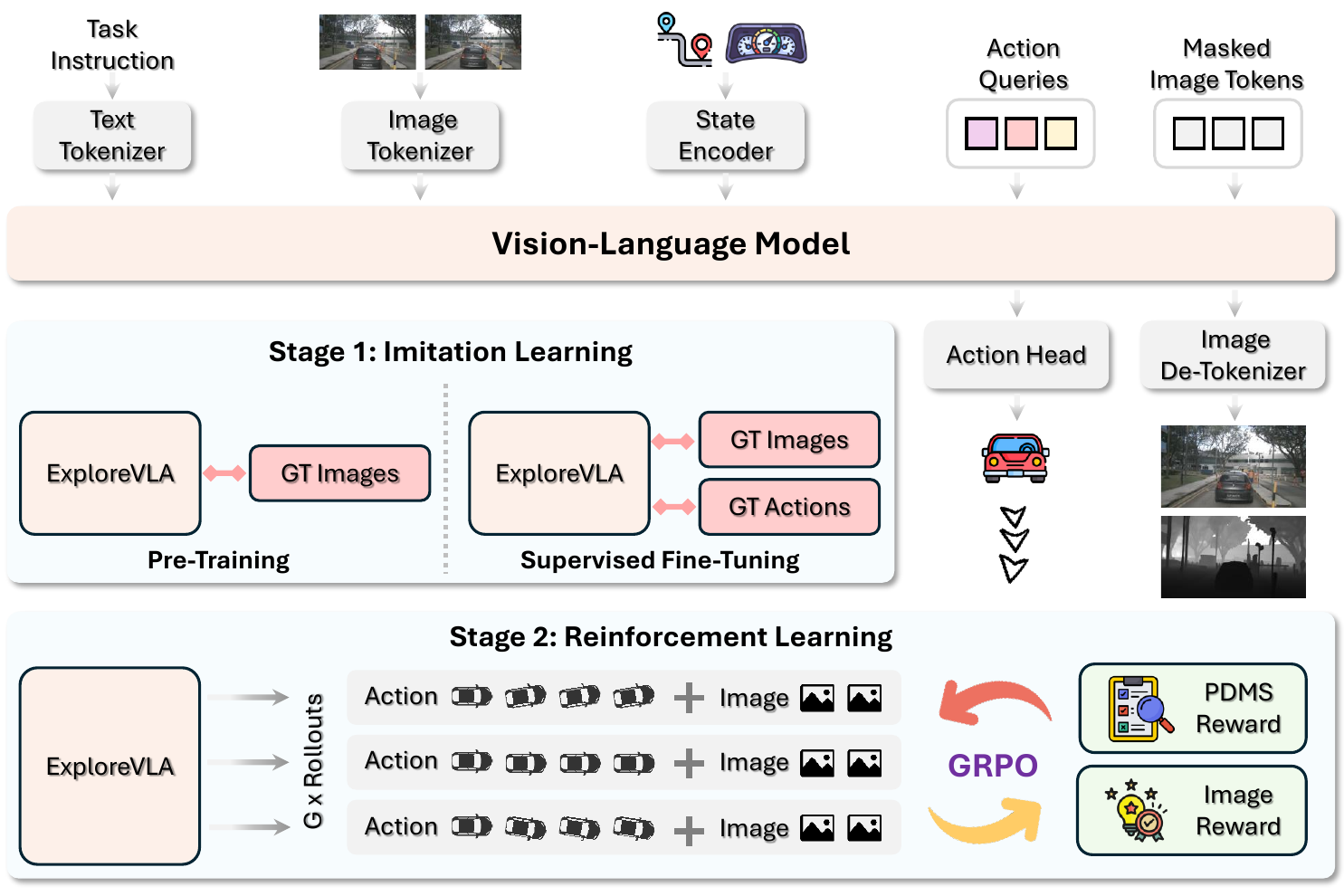}
  \caption{\textbf{Model architecture and training paradigm of ExploreVLA.} The model takes task instructions, multi-frame images, and ego status as input, and jointly predicts future trajectories and future images. Training proceeds in two stages: (1) imitation learning, consisting of pre-training on image generation and supervised fine-tuning on both actions and images, and (2) reinforcement learning, where GRPO optimizes the policy using a composite reward combining PDMS and image-based exploration bonus.}
  \label{fig:overview}
\end{figure}

We present a unified understanding-and-generation framework for end-to-end autonomous driving that addresses two key limitations of existing VLA models: (1) the lack of exploration beyond expert demonstrations and (2) the reliance on sparse supervisory signals. Our framework consists of three components. 
First, we build upon a unified VLM backbone that jointly supports autoregressive text modeling and discrete image generation within a single architecture (\cref{sec:architecture}). 
Second, we introduce future RGB and depth image generation as dense world modeling objectives that provide token-level supervision alongside trajectory prediction, encouraging the model to learn richer scene representations (\cref{sec:dense_supervision}). 
Third, we leverage the world model's prediction uncertainty as an intrinsic reward signal to guide policy exploration via Group Relative Policy Optimization (GRPO), enabling the model to discover diverse driving strategies beyond mere imitation (\cref{sec:exploration_reward}). An overview of our framework is illustrated in \cref{fig:overview}.

\subsection{Problem Formulation}
\label{sec:problem_formulation}

We formulate autonomous driving as a unified understanding-and-generation task, where a single model jointly predicts future trajectories and generates dense visual representations conditioned on the current driving context.

\subsubsection{Input.}
At each timestep $t$, the model receives three inputs: (1) the current and $T$ past front-view camera images $\{\mathbf{I}_{t-T}, \cdots, \mathbf{I}_{t}\}$ with $\mathbf{I} \in \mathbb{R}^{H \times W \times 3}$, (2) a natural language command $\mathbf{c}$, and (3) the ego-vehicle status $\mathbf{s}_t$.

\subsubsection{Output.}
The model produces three types of outputs:
(1) predicted future waypoints $\boldsymbol{\tau} = \{\hat{p}_i\}_{i=1}^{N_\tau}$, where $N_\tau$ denotes the planning horizon,
(2) predicted future RGB frames $\{\hat{\mathbf{I}}_{t+1}, \cdots, \hat{\mathbf{I}}_{t+F}\} $ with $\hat{\mathbf{I}} \in \mathbb{R}^{H' \times W' \times 3}$ capturing the anticipated visual appearance of the driving scene,
and (3) predicted future depth maps $\{\hat{\mathbf{D}}_{t+1}, \cdots, \hat{\mathbf{D}}_{t+F}\}$ with $\hat{\mathbf{D}} \in \mathbb{R}^{H' \times W'}$ encoding the geometric structure of the future scene. Here $F$ denotes the number of future frames to generate.

\subsection{Unified Understanding and Generation Architecture}
\label{sec:architecture}

% Our model builds upon the Show-o architecture~\cite{xie2024show}, a unified transformer that seamlessly combines autoregressive modeling and discrete diffusion within a single backbone. We adapt this framework to the autonomous driving setting, where the model must jointly predict future trajectories and generate dense visual representations of the future.

\subsubsection{Tokenization.}
We maintain a unified vocabulary of discrete tokens spanning text, images, and special task indicators. For \textbf{text tokenization}, we adopt the same tokenizer from the pre-trained LLM backbone, such that the language command $\mathbf{c}$ is tokenized into $L$ text tokens $\mathbf{v} = \{v_1, v_2, \cdots, v_L\}$. For \textbf{image tokenization}, we employ a pre-trained MAGVIT-v2~\cite{yu2023language} quantizer with a lookup-free codebook of size $K = 8{,}192$. Each image is encoded into discrete tokens by partitioning it into non-overlapping patches of size $16 \times 16$. Each of the $(T+1)$ input frames is independently tokenized into $M$ image tokens, yielding the input image token sequence $\mathbf{u} = \{u_1, u_2, \cdots, u_{(T+1) \times M}\}$. For \textbf{ego status encoding}, the $\mathbf{s}_t$ is projected into the transformer's embedding space via a learnable MLP, producing ego status embeddings $\mathbf{e}_s$ that are concatenated with the other input tokens.

\subsubsection{Causal and Full Attention Mechanism.}
We adopt the omni-attention mechanism from Show-o~\cite{xie2024show}, which adaptively combines causal and full attention depending on the token type. Text tokens $\mathbf{v}$ and ego status embeddings $\mathbf{e}_s$ are processed via \textit{causal attention}, where each token attends only to its preceding tokens. Image tokens $\mathbf{u}$ are processed via \textit{full attention}, which allows comprehensive interaction among all spatial positions and ensures that the generation process is fully conditioned on the driving context.
% For the future image tokens $\mathbf{u}^{\text{rgb}}_{\text{fut}}$ and $\mathbf{u}^{\text{dep}}_{\text{fut}}$, full attention is applied to leverage global context from unmasked tokens during generation. 
% Importantly, the future image tokens can attend to all preceding context, including historical frames, ego status, and the navigation command, which ensures that the generation process is fully conditioned on the driving context.

\subsubsection{Trajectory Prediction Head.}
In addition to the generation outputs, we extract the hidden states from the transformer at designated positions and feed them through a lightweight MLP head to predict future waypoints:
\begin{equation}
    \boldsymbol{\tau} = \text{MLP}(\mathbf{h}),
    \label{eq:traj_mlp}
\end{equation}
where $\mathbf{h}$ denotes the hidden representation. This design decouples trajectory prediction from the discrete token vocabulary, allowing the model to produce continuous-valued waypoints while sharing the same contextualized representations learned through the joint understanding-and-generation objective.

\subsection{Dense Supervisory Signals via World Modeling}
\label{sec:dense_supervision}

A central limitation of existing VLA models for autonomous driving is their reliance on sparse supervisory signals. Textual descriptions provide only high-level semantic intent (\eg, ``turn left''), while trajectory waypoints encode a thin, low-dimensional slice of the rich information embedded in driving scenes. As a result, the vast majority of scene structure, such as patial layout and depth ordering, remains unsupervised, which limits the model's ability to learn comprehensive representations of the driving environment.

We address this by formulating future scene generation as an auxiliary world modeling objective that provides dense supervision alongside trajectory prediction. Our model is trained to generate $F$ future frames of both RGB images and depth maps, effectively requiring it to ``imagine'' the future visual and geometric state of the world. This dense generation objective forces the model to internalize fine-grained knowledge about scene dynamics.

\subsubsection{RGB Generation as Visual Supervision.}
The future RGB generation objective requires the model to predict the visual appearance of upcoming driving scenes, providing dense supervision over texture, color, object identity, and scene semantics. Each future RGB frame is tokenized via the shared MAGVIT-v2 quantizer, and the model learns to reconstruct randomly masked tokens through the mask token prediction objective. Formally, the RGB generation loss over all $F$ future frames is:
\begin{equation}
    \mathcal{L}_{\text{rgb}} = -\sum_{f=1}^{F} \sum_{j} \log p_\theta \big( u^{\text{rgb}}_{f,j} \mid \mathbf{u}^{\text{rgb}}_{f,*},\ \mathbf{v},\ \mathbf{e}_s,\ \mathbf{u} \big),
    \label{eq:rgb_loss}
\end{equation}
where $u^{\text{rgb}}_{f,j}$ is the $j$-th masked token of the $f$-th future RGB frame, and $\mathbf{u}^{\text{rgb}}_{f,*}$ denotes the corresponding masked sequence. By reconstructing future visual tokens, the model learns to capture how scene appearance evolves under the current driving context.

\subsubsection{Depth Generation as Geometric Supervision.}
Complementary to RGB, the depth generation objective supervises the model on the 3D geometric structure of future scenes. Depth maps encode object distances, spatial layout, and surface orientation, which is critical for safe planning but entirely absent from text or trajectory supervision. The depth generation loss is defined analogously:
\begin{equation}
    \mathcal{L}_{\text{depth}} = -\sum_{f=1}^{F} \sum_{j} \log p_\theta \big( u^{\text{dep}}_{f,j} \mid \mathbf{u}^{\text{dep}}_{f,*},\ \mathbf{u}^{\text{rgb}}_{f,*},\ \mathbf{v},\ \mathbf{e}_s,\ \mathbf{u} \big),
    \label{eq:depth_loss}
\end{equation}
where $u^{\text{dep}}_{f,j}$ is the $j$-th masked token of the $f$-th future depth map. The overall mask token prediction (MTP) loss decomposes as:
\begin{equation}
    \mathcal{L}_{\text{MTP}} = \mathcal{L}_{\text{rgb}} + \mathcal{L}_{\text{depth}}.
    \label{eq:mtp_decompose}
\end{equation}
% Together with $\mathcal{L}_{\text{NTP}}$ and $\mathcal{L}_{\text{traj}}$, the full first-stage training objective follows \cref{eq:total_loss}. 
% By jointly optimizing across these objectives, the model leverages dense visual and geometric supervision to learn representations that are substantially richer than those obtained from sparse trajectory-only training, ultimately translating into more accurate and robust planning.

\subsection{Intrinsic Reward from World Model for Exploration}
\label{sec:exploration_reward}

Models trained purely via behavior cloning tend to narrowly replicate the expert's actions and struggle to generalize when the test-time distribution deviates from the training data. In the second stage of training, we leverage the world model learned in the first stage as a source of intrinsic reward signals that encourage the policy to explore novel yet safe driving behaviors beyond mere imitation. The core idea is that the world model's image prediction uncertainty provides a natural measure of trajectory novelty relative to the entire training distribution: high uncertainty indicates that the model has rarely encountered the visual consequences of a given action, signaling an out-of-distribution trajectory that, if safe, represents a valuable learning opportunity.

\subsubsection{Uncertainty-Based Exploration Bonus.}
Given a candidate trajectory $\boldsymbol{\tau}_i$ sampled from the current VLA policy, we condition the world model on $\boldsymbol{\tau}_i$ and perform future image generation. For each predicted discrete image token, the model outputs a probability distribution over the MAGVIT-v2 codebook. We quantify the prediction uncertainty using the entropy of these token-level distributions, averaged across all generated future RGB and depth tokens:
\begin{equation}
    \mathcal{H}(\boldsymbol{\tau}_i) = -\frac{1}{|\mathcal{M}|}\sum_{j \in \mathcal{M}} p_j \log p_j,
    \label{eq:entropy}
\end{equation}
where $\mathcal{M}$ denotes the set of generated image token positions across all future RGB and depth frames, and $p_j$ is the predicted probability of image token $j$. A high entropy indicates that the world model is uncertain about the visual consequences of trajectory $\boldsymbol{\tau}_i$, meaning the trajectory leads to scenarios underrepresented in the training distribution. Conversely, low entropy indicates an in-distribution trajectory whose consequences the model has already learned to predict. We define the exploration bonus as the normalized entropy:
\begin{equation}
    b_i = f(\mathcal{H}(\boldsymbol{\tau}_i)),
    \label{eq:exploration_bonus}
\end{equation}
where $f(\mathcal{H}) = (\mathcal{H}-\mathcal{H}_{\min})/(\mathcal{H}_{\max}-\mathcal{H}_{\min})$ maps the raw entropy to a normalized bonus $b_i \in [0, 1]$.

\subsubsection{Safety-Gated Reward.}
High uncertainty alone does not guarantee that exploration is beneficial, since a trajectory leading to a collision is novel but not useful. We therefore gate the exploration bonus using the PDMS~\cite{dauner2024navsim}, which evaluates trajectory quality on a $[0, 1]$ scale based on collision avoidance, comfort, and progress. The intrinsic reward for trajectory $\boldsymbol{\tau}_i$ is:
\begin{equation}
    R_i = 
    \begin{cases}
        \text{PDMS}_i + \lambda \cdot b_i, & \text{if } \text{PDMS}_i > \delta, \\
        \text{PDMS}_i, & \text{otherwise},
    \end{cases}
    \label{eq:reward}
\end{equation}
where $\delta$ is a safety threshold and $\lambda$ controls the exploration strength. The gating mechanism ensures that only safe trajectories ($\text{PDMS}_i > \delta$) receive the exploration bonus. Unsafe or failed explorations are scored purely by their PDMS without additional encouragement. This design prioritizes trajectories that are simultaneously novel (high world model uncertainty) and good (high PDMS). Such trajectories represent out-of-distribution behaviors that are critical for improving generalization beyond expert demonstrations.

\subsubsection{Policy Optimization via GRPO.}
We optimize the policy using GRPO~\cite{shao2024deepseekmath}. At each iteration, we sample a group of $G$ candidate trajectories $\{\boldsymbol{\tau}_1, \cdots, \boldsymbol{\tau}_G\}$ from the current policy. Each trajectory and corresponding predicted image tokens are scored using the reward in \cref{eq:reward}, and rewards are normalized within the group to compute relative advantages:
\begin{equation}
    \hat{A}_i = \frac{R_i - \text{mean}(\{R_1, \cdots, R_G\})}{\text{std}(\{R_1, \cdots, R_G\})}.
    \label{eq:advantage}
\end{equation}
The policy is updated to increase the likelihood of trajectories with positive advantages and suppress those with negative advantages. This group-relative formulation naturally steers the policy toward trajectories that are both good and novel within each sampled group.

\subsection{Training Strategy}
As shown in \cref{fig:overview}, our training proceeds in two stages. The first stage consists of two phases: pre-training and supervised fine-tuning. During pre-training, ground-truth future actions are provided as input, and the model is trained solely with the image generation objective. This allows the model to adapt its visual generation capabilities to driving scenes. 
In the supervised fine-tuning phase, the model is trained to jointly predict both actions and future images. 
In the second stage, we further refine the policy using an intrinsic reward derived from the world model to encourage exploration.

\section{Experiments}

\subsection{Experimental Setup}

\subsubsection{Datasets.}
We evaluate our method on two widely used autonomous driving benchmarks: NAVSIM and nuScenes.

\textbf{NAVSIM}~\cite{dauner2024navsim,cao2025pseudo} is a recently proposed non-reactive simulation benchmark built upon the OpenScene dataset. NAVSIM v1 evaluates planning quality using PDMS~\cite{dauner2024navsim}, and NAVSIM v2 adopts the Extended PDMS (EPDMS)~\cite{cao2025pseudo}. Both metrics aggregate factors such as progress, time-to-collision, and comfort, achieving stronger correlation with closed-loop evaluations. We train on the \texttt{navtrain} split and report results on the \texttt{navtest} split. In NAVSIM, the model predicts future waypoints in the form of $(x, y, \theta)$ over a 4-second planning horizon.

\textbf{nuScenes}~\cite{caesar2020nuscenes} consists of 1,000 driving scenes and has become a standard benchmark for open-loop planning evaluation. Following prior works~\cite{hu2022st,hu2023planning,jiang2023vad}, we adopt the standard train/val split and evaluate using the L2 displacement error and collision rate between predicted and ground-truth trajectories. On nuScenes, the model predicts future waypoints in the form of $(x, y)$. The evaluation results on nuScenes are presented in the supplementary material.

\subsubsection{Implementation Details.}
Our model is built upon Show-o~\cite{xie2024show} as the backbone architecture. In the first training stage, we first pre-train the model for 10 epochs by conditioning on ground-truth future actions and supervising only the image generation. We then perform supervised fine-tuning for 15 epochs, where the model jointly predicts future trajectories and generates future RGB and depth images. In the second stage, we apply GRPO-based post-training with LoRA~\cite{hu2022lora} for 5 epochs to further refine the policy using the exploration-aware reward described in \cref{sec:exploration_reward}. All experiments are conducted on 4$\times$H200 GPUs. We obtain depth maps from a pre-trained monocular depth estimation model~\cite{yin2023metric3d}. Detailed hyperparameters are provided in the supplementary material.

\begin{table}[tb]
  \caption{\textbf{Comparison on NAVSIM v1 with closed-loop metrics.} The best performance is marked in \textbf{bold}, and the second best is \underline{underlined}. 
  Abbreviations: no at-fault collision (NC), drivable area compliance (DAC), ego progress (EP), time to collision (TTC), comfort (Comf.).
  SC: single-view camera; MC: multi-view camera; L: LiDAR; $\dagger$: best-of-N (N=6) strategy~\cite{zhou2025autovla}. 
  }
  \label{tab:navsim_v1}
  \centering
  \resizebox{\textwidth}{!}{
  \setlength{\tabcolsep}{4pt}
  \begin{tabular}{l|c|ccccc|>{\columncolor{gray!25}}c}
    \toprule
    Model                                      & Input & NC $\uparrow$ & DAC $\uparrow$  & EP $\uparrow$ & TTC $\uparrow$  & Comf. $\uparrow$ & PDMS $\uparrow$ \\
    \midrule
    Ego Status MLP                              & -    & 93.1   & 78.3          & 63.2           & 84.0  & \underline{99.9}            & 66.4   \\
    TransFuser~\cite{chitta2022transfuser}      & -    & 97.8   & 92.6          & 78.9           & 92.9  & \underline{99.9}            & 83.9   \\
    DRAMA~\cite{yuan2024drama}                  & MC+L & 98.2   & 95.2          & 81.3           & 94.2  & \textbf{100.0}  & 86.9   \\
    Hydra-MDP~\cite{li2024hydra}                & MC+L & 99.1   & 98.3          & 85.2           & 96.6  & \textbf{100.0}  & 91.3   \\
    Centaur~\cite{sima2025centaur}              &  MC+L & 99.2   & 98.7          & 86.0           & 97.2  & \underline{99.9}            & 92.1   \\
    DriveSuprim~\cite{yao2025drivesuprim}       &  MC+L & 98.6   & 98.6          & \textbf{91.3}  & 95.5  & \textbf{100.0}            & \underline{93.5}   \\ 
    \midrule
    DrivingGPT~\cite{chen2025drivinggpt}        & SC   & 98.9   & 90.7          & 79.9           & 94.9  & 95.6            & 82.4   \\
    FSDrive~\cite{zeng2025futuresightdrive}     & SC   & 98.2   & 93.8          & 80.1           & 93.3  & \underline{99.9}            & 85.1   \\
    PWM~\cite{zhao2025pwm}                      & SC   & 98.9   & 95.8          & 81.5           & 95.9  & \textbf{100.0}  & 88.1   \\
    AutoVLA~\cite{zhou2025autovla}              & MC   & 98.4   & 95.6          & 81.9           & \underline{98.0}  & \underline{99.9}            & 89.1   \\
    AutoVLA$\dagger$~\cite{zhou2025autovla}    & MC   & 99.1   & 97.1          & 87.6           & 97.1  & \underline{99.9}            & 92.1   \\
    DriveVLA-W0~\cite{li2025drivevla}           & SC   & 98.7   & \textbf{99.1} & 87.6           & 97.1  & \textbf{100.0}  & 90.2   \\
    DriveVLA-W0$\dagger$~\cite{li2025drivevla}  & SC   & \textbf{99.9}   & 97.4          & \underline{88.3}           & 97.0  & \underline{99.9}            & 93.0   \\
    \midrule
    ExploreVLA                                  & SC   & 98.8   & 98.4          & 83.5           & 96.5  & \underline{99.9}            & 90.4   \\
    ExploreVLA$\dagger$                        & SC   & \underline{99.4}  & \underline{98.9}  & \underline{88.3}  & \textbf{98.3}   & 99.7  & \textbf{93.7}  \\
    \bottomrule
  \end{tabular}
  }
\end{table}

\begin{table}[tb]
\centering
\caption{\textbf{Comparison on NAVSIM v2 with extended closed-loop metrics.} The best performance is marked in \textbf{bold}, and the second best is \underline{underlined}.}
\label{tab:navsim_v2}
\resizebox{\textwidth}{!}{
\begin{tabular}{l|cccc|ccccc|>{\columncolor{gray!25}}c}
\toprule
\textbf{Model} & NC $\uparrow$ & DAC $\uparrow$ & DDC $\uparrow$ & TLC $\uparrow$ & EP $\uparrow$ & TTC $\uparrow$ & LK $\uparrow$ & HC $\uparrow$ & EC $\uparrow$ & EPDMS $\uparrow$ \\
\midrule
Ego Status & 93.1 & 77.9 & 92.7 & 99.6 & 86.0 & 91.5 & 89.4 & \textbf{98.3} & 85.4 & 64.0 \\
TransFuser~\cite{chitta2022transfuser} & 96.9 & 89.9 & 97.8 & \underline{99.7} & 87.1 & 95.4 & 92.7 & \textbf{98.3} & 87.2 & 76.7 \\
Hydra-MDP++~\cite{li2025hydra} & 97.2 & 97.5 & \underline{99.4} & 99.6 & 83.1 & 96.5 & 94.4 & \underline{98.2} & 70.9 & 81.4 \\
DriveSuprem~\cite{yao2025drivesuprim} & 97.5 & 96.5 & \underline{99.4} & 99.6 & \underline{88.4} & 96.6 & 95.5 & \textbf{98.3} & 77.0 & 83.1 \\
ARTEMIS~\cite{feng2025artemis} & 98.3 & 95.1 & 98.6 & \textbf{99.8} & 81.5 & 97.4 & 96.5 & \textbf{98.3} & - & 83.1 \\
DiffusionDrive~\cite{liao2025diffusiondrive} & 98.2 & 95.9 & \underline{99.4} & \textbf{99.8} & 87.5 & 97.3 & \underline{96.8} & \textbf{98.3} & \underline{87.7} & 84.5 \\ 
DiffusionDriveV2~\cite{zou2025diffusiondrivev2} & 97.7 & \underline{96.6} & 99.2 & \textbf{99.8} & \textbf{88.9} & 97.2 & 96.0 & 97.8 & \textbf{91.0} & 85.5 \\ 
DriveVLA-W0~\cite{li2025drivevla} & \underline{98.5} & \textbf{99.1} & 98.0 & \underline{99.7} & 86.4 & \underline{98.1} & 93.2 & 97.9 & 58.9 & \underline{86.1} \\
\midrule
ExploreVLA & \textbf{98.8} & 96.2 & \textbf{99.6} & \textbf{99.8} & 87.1 & \textbf{98.2} & \textbf{97.8} & \textbf{98.3} & 86.8 & \textbf{88.8} \\
\bottomrule
\end{tabular}
}
\end{table}

\subsection{Main Results}
\subsubsection{Results on NAVSIM v1.}
\cref{tab:navsim_v1} presents the comparison on the NAVSIM v1 benchmark. Our ExploreVLA achieves the highest PDMS of 93.7 with the best-of-N strategy, outperforming all prior approaches. Notably, ExploreVLA uses only a single-view camera, yet surpasses multi-sensor methods such as DriveSuprim and Centaur. Even without the best-of-N strategy, ExploreVLA attains a PDMS of 90.4, which is competitive with multi-view methods like AutoVLA. Among the sub-metrics, ExploreVLA$\dagger$ achieves the best TTC and the second-best scores on NC, DAC, and EP, demonstrating that our world-model-based exploration reward helps the model internalize diverse and robust driving behaviors beyond what pure imitation learning can provide.
% consistently improves safety and progress rather than optimizing a single dimension.

\subsubsection{Results on NAVSIM v2.}
\cref{tab:navsim_v2} further validates our approach on NAVSIM v2, which introduces extended closed-loop metrics including driving direction compliance (DDC), traffic light compliance (TLC), lane keeping (LK), history comfort (HC), and extended comfort (EC). ExploreVLA achieves the highest EPDMS of 88.8, surpassing the previous best result of 86.1 by DriveVLA-W0 by 2.7 points. Our method obtains the best scores on six out of nine individual metrics, while remaining highly competitive on the rest.

\begin{figure}[tb]
  \centering
  \includegraphics[width=0.99\textwidth]{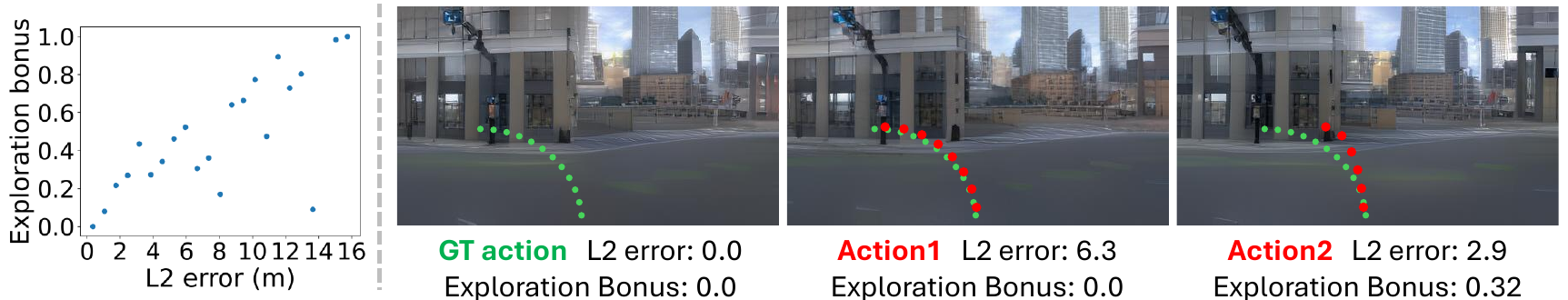}
  \caption{\textbf{Analysis of the exploration bonus.} Left: the exploration bonus is positively correlated with L2 error to the ground-truth trajectory. Right: our exploration bonus can properly measure the trajectory novelty that L2 error fails.}
  \label{fig:reward}
\end{figure}

\subsection{Analysis of Intrinsic Reward Modeling}

\cref{fig:reward} provides an analysis of our intrinsic reward modeling mechanism. The left shows a generally positive correlation between the exploration bonus and L2 error with respect to the ground-truth trajectory: as the sampled trajectory deviates further from the expert action, the world model's prediction uncertainty tends to increase, resulting in higher exploration bonuses. 
However, L2 distance is not always an unreliable measure of novelty, and our uncertainty-based bonus reflects it more faithfully. The right of \cref{fig:reward} shows a representative failure case: a trajectory that closely follows the expert's direction may incur a large L2 error due to positional shift, while a trajectory that takes a fundamentally different route may have a smaller L2 error. 
{Additionally, we randomly select 1{,}000 straight-driving scenes and apply two types of perturbation to the expert trajectory: (1) speed variation, which changes vehicle speed while following nearly identical paths, and (2) direction variation, which changes the driving direction and therefore represents genuinely different driving behaviors. As shown in \cref{tab:l2_vs_bonus}, the speed perturbation produces a substantially larger L2 error despite exhibiting little behavioral novelty, whereas the direction perturbation yields a smaller L2 error while receiving a significantly higher exploration bonus. These results indicate that the proposed uncertainty-based reward better captures behavioral novelty than trajectory distance.}

% {To further understand the role of the proposed intrinsic reward, we analyze the relationship between the world model prediction entropy and the PDMS score in right of Fig.~\ref{fig:reward}. The distribution separates into four interpretable regions. Most trajectories exhibit both low entropy and high PDMS, corresponding to safe in-distribution behaviors that closely resemble expert demonstrations. More importantly, our method identifies trajectories with simultaneously high entropy and high PDMS, which represent novel yet safe driving strategies that should be encouraged during exploration. In contrast, trajectories with high entropy but low PDMS correspond to novel yet unsafe behaviors. This observation justifies the necessity of the safety-gated reward in Eq.~(\ref{eq:reward}), which selectively encourages safe exploratory behaviors while suppressing unsafe ones. Overall, these results demonstrate that entropy and PDMS capture complementary information: entropy measures behavioral novelty with respect to the training distribution, whereas PDMS evaluates driving quality.}

\begin{table}[t]
  \centering
  \caption{\textbf{L2 error vs.\ exploration bonus on 1{,}000 randomly selected straight-driving scenes under two perturbation types.} L2 error misjudges novelty in both cases, whereas our exploration bonus correctly assigns low novelty to speed changes (similar path) and high novelty to direction changes (different path).}
  \label{tab:l2_vs_bonus}
  \resizebox{\textwidth}{!}{
  \setlength{\tabcolsep}{4pt}
  \begin{tabular}{lcc}
    \toprule
     & Type~1 (speed $\Delta$, similar path) & Type~2 (direction $\Delta$, different path) \\
    \midrule
    L2 (m)  & $5.82 \pm 1.74$ & $2.15 \pm 0.93$ \\
    Exploration bonus & $0.04 \pm 0.03$ & $0.29 \pm 0.12$ \\
    \bottomrule
  \end{tabular}
  }
\end{table}

\subsection{Ablation Study}

\begin{table}[tb]
  \caption{\textbf{Ablation study on dense visual supervision.} We evaluate the effect of auxiliary RGB and depth image generation during Stage 1 imitation learning on the NAVSIM v1 \texttt{navtest} split.}
  \label{tab:ablation_gen}
  \centering
  % \resizebox{0.75\textwidth}{!}{
  \setlength{\tabcolsep}{4pt}
  \begin{tabular}{cc|ccccc|c}
    \toprule
    \makecell{RGB Img. \\ Gen.} & \makecell{Depth Img. \\ Gen.} & NC $\uparrow$ & DAC $\uparrow$  & EP $\uparrow$ & TTC $\uparrow$  & Comf. $\uparrow$ & PDMS $\uparrow$ \\
    \midrule
    \ding{55} & \ding{55} & \underline{98.6} & 94.4 & 80.1 & 94.8 & \textbf{100.0} & 86.2 \\
    \ding{51} & \ding{55} & \textbf{98.7} & \underline{96.0} & \underline{81.5} & 95.4 & \underline{99.9} & \underline{87.9} \\
    \ding{55} & \ding{51} & \textbf{98.7} &  95.8 & 81.4 & \underline{95.6} & \underline{99.9} & 87.8 \\
    \rowcolor{blue!15}
    \ding{51} & \ding{51} & \textbf{98.7} & \textbf{96.3} & \textbf{82.3} & \textbf{95.7} & \underline{99.9} & \textbf{88.5} \\
    \bottomrule
  \end{tabular}
  % }
\end{table}

\subsubsection{Effect of Dense Visual Supervision.}
\cref{tab:ablation_gen} examines the contribution of RGB and depth image generation as auxiliary supervision during Stage 1 training. The trajectory-only baseline without any image generation achieves the lowest PDMS. Adding either RGB or depth generation alone yields comparable improvements, which confirms that both modalities provide meaningful dense supervisory signals for learning better scene representations. Combining both RGB and depth generation further improves PDMS to 88.5. This indicates that RGB and depth capture complementary information (visual appearance and geometric structure) and their joint supervision leads to a more comprehensive understanding of driving scenes.

\subsubsection{Effect of Reward Design.}
\cref{tab:ablation_reward} isolates the contribution of each reward component during Stage 2 RL. Starting from the Stage 1 model (PDMS 88.50), applying only the PDMS reward brings a substantial improvement to 90.19, demonstrating the effectiveness of GRPO-based post-training. Using the image-based exploration reward alone yields only a marginal gain (88.53), which is expected since the image reward serves as an exploration bonus rather than a direct driving quality signal. Without the safety gate provided by PDMS thresholding, the exploration signal alone cannot effectively guide policy improvement. When combining both rewards, the model achieves the best PDMS of 90.36. This confirms that the image-based exploration reward provides a complementary learning signal that encourages the policy to discover diverse driving strategies beyond what the PDMS reward alone can incentivize.

\begin{table}[tb]
  \caption{\textbf{Ablation study on reward components.} We evaluate the contribution of each reward signal during Stage 2 reinforcement learning on the NAVSIM v1 \texttt{navtest} split. The first row (no reward) corresponds to the Stage 1 baseline.}
  \label{tab:ablation_reward}
  \centering
  % \resizebox{0.75\textwidth}{!}{
  \setlength{\tabcolsep}{4pt}
  \begin{tabular}{cc|ccccc|c}
    \toprule
    \makecell{PDMS \\ Reward} & \makecell{Image \\ Reward} & NC $\uparrow$ & DAC $\uparrow$  & EP $\uparrow$ & TTC $\uparrow$  & Comf. $\uparrow$ & PDMS $\uparrow$ \\
    \midrule
    \ding{55} & \ding{55} & 98.74 & 96.25 & 82.27 & 95.67 & \textbf{99.97} & 88.50 \\
    \ding{51} & \ding{55} & \textbf{98.82} & \underline{97.78} & \underline{83.44} & \underline{96.50} & \underline{99.95} & \underline{90.19} \\
    \ding{55} & \ding{51} & 98.76 & 96.30 & 82.32 & 95.65 & \textbf{99.97} & 88.53 \\
    \rowcolor{blue!15}
    \ding{51} & \ding{51} & \underline{98.81} & \textbf{97.88} & \textbf{83.53} & \textbf{96.66} & \underline{99.95} & \textbf{90.36}   \\
    \bottomrule
  \end{tabular}
  % }
\end{table}

\begin{table}[tb]
\caption{\textbf{Ablation on the safety threshold $\delta$.} We evaluate the effect of different safety thresholds in \cref{eq:reward} during Stage 2  reinforcement learning on the NAVSIM v1 \texttt{navtest} split.}
\label{tab:delta}
  \centering
  % \resizebox{0.75\textwidth}{!}{
  \setlength{\tabcolsep}{4pt}
  \begin{tabular}{c|ccccc|c}
    \toprule
    $\delta$ & NC $\uparrow$ & DAC $\uparrow$  & EP $\uparrow$ & TTC $\uparrow$  & Comf. $\uparrow$ & PDMS $\uparrow$ \\
    \midrule
    0.0 & \textbf{98.85} & \underline{97.82} & \underline{83.48} & 96.53 & 99.92 & \underline{90.25} \\
    0.3 & \underline{98.83} & 97.78 & 83.47 & 96.49 & 99.91 & 90.21 \\
    0.6 & 98.81 & 97.79 & 83.44 & \underline{96.54} & \underline{99.93} & 90.22 \\
    \rowcolor{blue!15}
    0.9 & 98.81 & \textbf{97.88} & \textbf{83.53} & \textbf{96.66} & \textbf{99.95} & \textbf{90.36}  \\
    \bottomrule
  \end{tabular}
  % }
\end{table}

\subsubsection{Effect of the Safety Threshold.}
The safety threshold $\delta$ in \cref{eq:reward} controls which trajectories are eligible for the exploration bonus. We ablate $\delta \in \{0, 0.3, 0.6, 0.9\}$ in \cref{tab:delta}. The proposed exploration reward consistently improves over the PDMS-only baseline (second row in \cref{tab:ablation_reward}) across all threshold choices. This result indicates that the effectiveness of our method is not sensitive to the exact value of $\delta$. Among all settings, $\delta=0.9$ achieves the best performance. The similar performance observed for $\delta \in \{0,0.3,0.6\}$ can be explained by the Stage 1 policy distribution. Approximately $99\%$ of the sampled trajectories already achieve PDMS scores above $0.6$, and thus making these thresholds insufficient for distinguishing safe trajectories from risky ones. When $\delta$ is increased to $0.9$, the safety gate begins to effectively separate safe-and-novel trajectories from unsafe exploratory behaviors, leading to the highest PDMS.

\subsubsection{Qualitative Analysis.}

\begin{figure}[tb]
  \centering
  \includegraphics[width=0.8\textwidth]{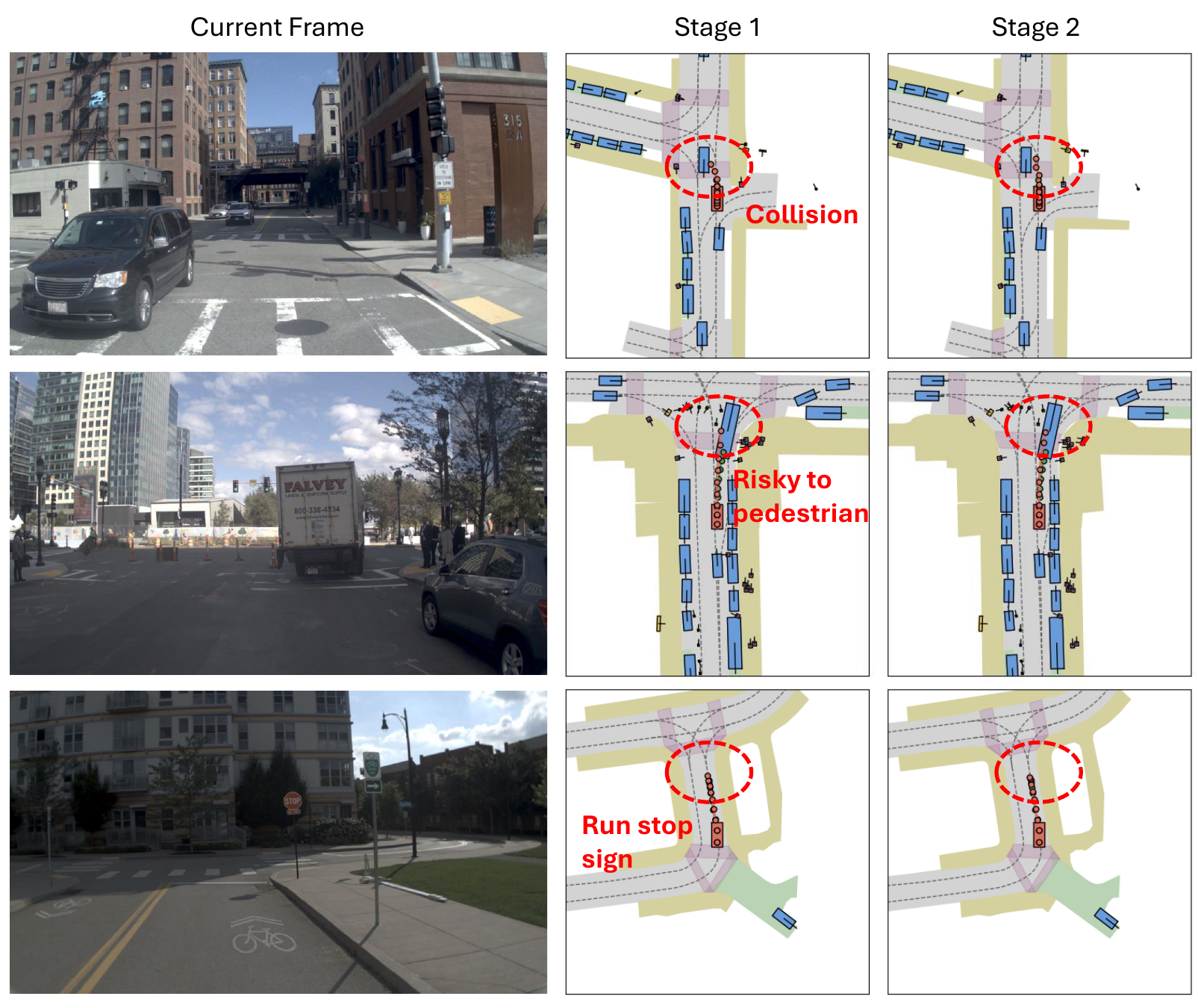}
  \caption{\textbf{Qualitative comparison of planned trajectories before and after RL post-training.} We visualize three challenging driving scenarios in bird's-eye view. The Stage 1 model exhibits safety-critical failures. After Stage 2 RL post-training, the model produces safer and more compliant trajectories. green: GT, orange: prediction.
  }
  \label{fig:qualitative}
\end{figure}

% \cref{fig:qualitative} visualizes the planned trajectories before (Stage 1) and after (Stage 2) RL in three representative scenarios. In the first example, the Stage 1 model plans a trajectory that collides with a vehicle at an intersection, whereas the Stage 2 model adjusts its path to maintain a safe distance. In the second example, the Stage 1 model generates a trajectory that passes dangerously close to pedestrians; after RL post-training, the model produces a more conservative trajectory that avoids the risky region. In the third example, the Stage 1 model fails to comply with a stop sign and continues through the intersection, while the Stage 2 model correctly yields. 
% These cases illustrate that our world-model-based RL not only improves aggregate metrics but also rectifies safety-critical failures that pure imitation learning struggles to resolve.
% from expert demonstrations alone.

\cref{fig:qualitative} visualizes planned trajectories after Stage 1 and Stage 2 in three representative scenarios. In each case, the Stage 1 model exhibits a safety-critical failure (\ie, colliding with a vehicle, passing dangerously close to pedestrians, or running a stop sign). After Stage 2 reinforcement learning, the model successfully corrects these behaviors. These results demonstrate that our world-model-based RL not only improves aggregate metrics but also rectifies safety-critical failures that pure imitation learning struggles to resolve from expert demonstrations alone.

\section{Conclusion}

We presented ExploreVLA, a unified framework that addresses the lack of exploration and sparse supervision in VLA-based autonomous driving. By jointly predicting future trajectories, RGB images, and depth maps, our approach provides dense world modeling supervision that enriches scene representations. We further leverage the world model's prediction uncertainty as an intrinsic novelty measure, combined with a safety-gated reward and GRPO, to guide the policy toward diverse yet safe driving strategies beyond expert imitation. 
Extensive experiments on the NAVSIM and nuScenes benchmarks validate the effectiveness of our approach, achieving state-of-the-art performance on NAVSIM.

\textbf{Limitations and Future Work.} 
Our framework currently uses a single front-view camera; extending to multi-view inputs could further broaden spatial coverage and planning robustness. 
% Additionally, exploring complementary generation targets such as Bird's-Eye View layouts may offer additional supervisory signals for scene structure, potentially further enhancing planning performance.
Moreover, our reinforcement learning stage relies on offline/open-loop post-training, which may bias the policy toward the training distribution and provide limited opportunities to learn recovery behaviors from self-induced states. Future work will investigate integrating our uncertainty-guided exploration framework with large-scale closed-loop reinforcement learning and evaluation.

% \clearpage\mbox{}Page \thepage\ of the manuscript. This is the last page.
% \par\vfill\par
% Now we have reached the maximum length of an ECCV \ECCVyear{} submission (excluding references and acknowledgements).
% References should start immediately after the main text, but can continue past p.\ 14 if needed. 
% \clearpage  % TODO FINAL: This \clearpage needs to be removed from both review and camera-ready versions.

% \section*{Acknowledgements}
% {Please insert your acknowledgments here.}

% ---- Bibliography ----
%
% BibTeX users should specify bibliography style 'splncs04'.
% References will then be sorted and formatted in the correct style.
%
\bibliographystyle{splncs04}
\bibliography{main}

\appendix

\section{More Implementation Details}

Our model is built upon Show-o~\cite{xie2024show}, which employs Phi-1.5~\cite{li2023textbooks} as the LLM backbone and a pre-trained MAGVIT-v2~\cite{yu2023language} as the image tokenizer. The input to the model consists of the current front-view image and one historical image captured 0.5 seconds earlier, and the model predicts the future RGB and depth images 0.5 seconds ahead. Input images are resized to $256 \times 448$. The ground-truth depth maps used for supervision are generated by Metric3D-ViT-Giant2~\cite{yin2023metric3d}. 

We use AdamW as the optimizer across all training stages. In Stage 1, the learning rate is $3 \times 10^{-5}$. In Stage 2, we apply LoRA~\cite{hu2022lora} with rank 32 and train with a learning rate of $3 \times 10^{-6}$. For GRPO, the group size is set to $G=8$, with KL penalty coefficient $\beta = 0.01$ and clipping range $\epsilon = 0.1$. The safety threshold $\delta$ in Eq.~(7) is set to 0.9, and the exploration bonus weight $\lambda$ is set to 0.5. The per-GPU batch size is 8 in Stage 1 and 1 in Stage 2. All experiments are conducted on 4$\times$H200 GPUs.

\section{More Experimental Results}
\subsection{Evaluation on NAVSIM}
\subsubsection{More Qualitative Results.}
\begin{figure}
  \centering
  \includegraphics[width=0.99\textwidth]{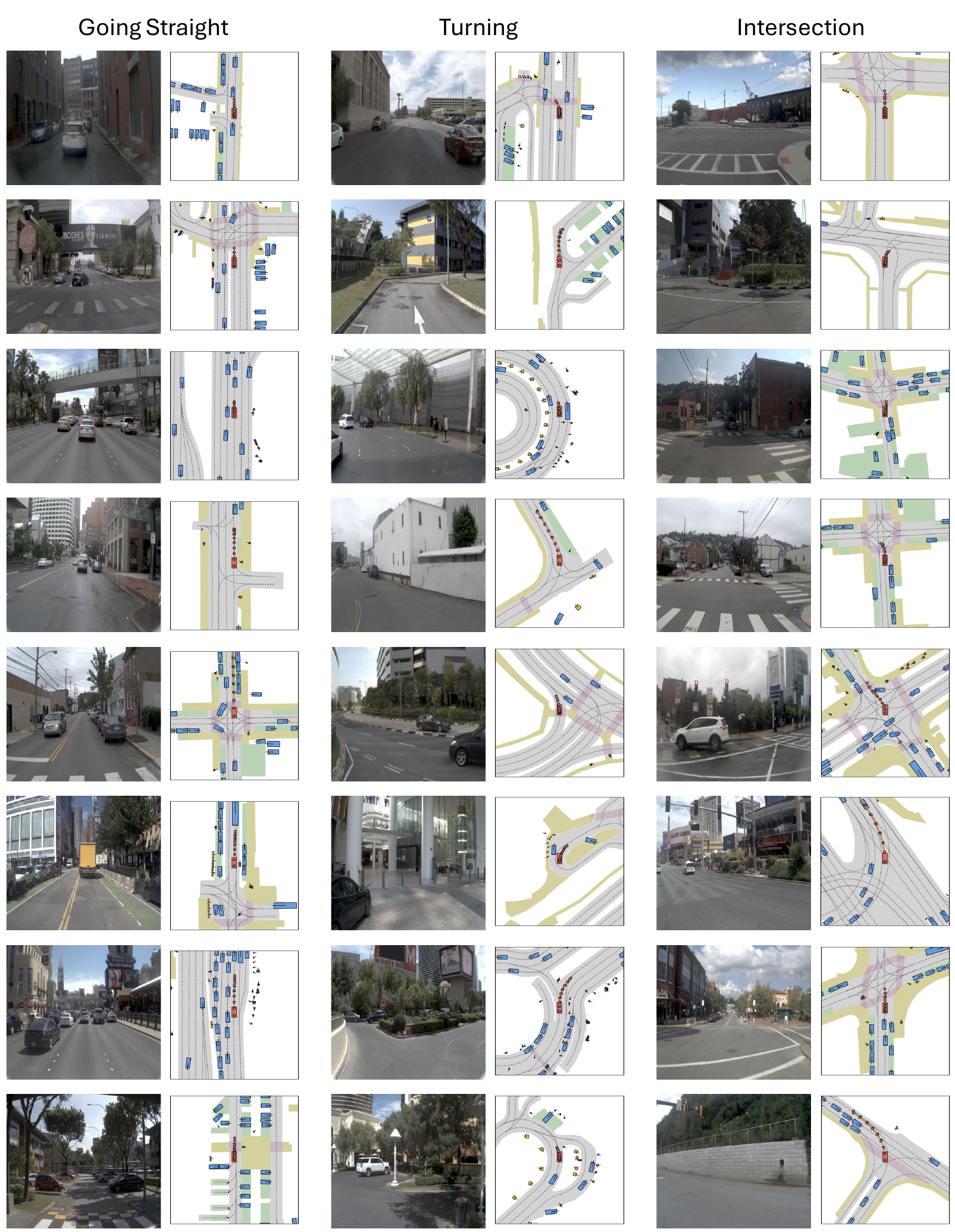}
  \caption{\textbf{Additional qualitative results on the \texttt{navtest} split.} We visualize the planned trajectories across three scenario categories: \textit{Going Straight}, \textit{Turning}, and \textit{Intersection}. For each example, we show the front-view camera image and the corresponding BEV representation with trajectories overlaid (green: GT, orange: prediction).}
  \label{fig:more_qualitative}
\end{figure}

We present additional qualitative results on the \texttt{navtest} split in \cref{fig:more_qualitative}, covering three representative scenario categories: Going Straight, Turning, and Intersection. Across all scenarios, our method generates planned trajectories that closely align with the road topology and traffic context. 
In straight-driving cases, the predicted trajectories maintain stable lane-keeping behavior. For turning scenarios, our model produces smooth and well-timed turning maneuvers that conform to the curvature of the road. Notably, in complex intersection scenarios involving multiple lanes and diverse traffic participants, our method still plans reasonable and safe trajectories, demonstrating its ability to handle intricate spatial layouts and dynamic interactions. 
% These results further validate the effectiveness of our dense world modeling supervision and novelty-aware exploration in learning robust driving policies that generalize across diverse real-world conditions.

\subsubsection{Image Generation from Dense World Modeling}
\begin{figure}
  \centering
  \includegraphics[width=0.99\textwidth]{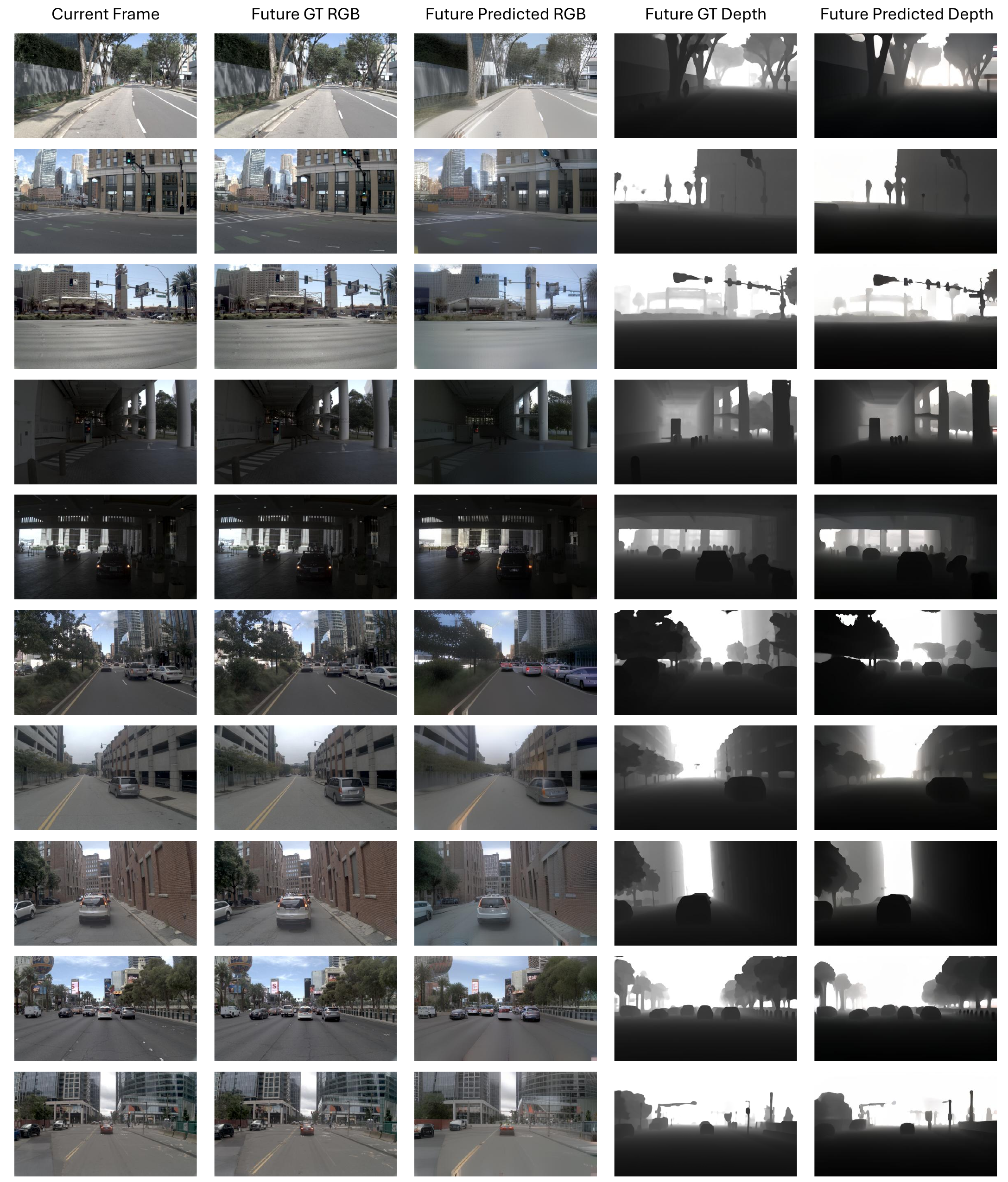}
  \caption{\textbf{Qualitative results of dense world modeling on the \texttt{navtest} split.} Each row shows a driving scenario with five columns: the current frame, ground truth and predicted future RGB images, and ground truth and predicted future depth maps.}
  \label{fig:vis_generation}
\end{figure}

\cref{fig:vis_generation} presents qualitative results of our dense world modeling on the \texttt{navtest} split. Given the current and historical frames, our model generates future RGB images and depth maps that closely align with the ground truth. 
We note that while fine-grained details such as texture sharpness and thin structures (\eg, traffic lights, tree branches) exhibit some degradation compared to ground truth, the global scene geometry and layout are well preserved. 
% These results confirm that our dense world modeling objective provides a rich supervisory signal that encodes both appearance and structural understanding, which in turn benefits the downstream planning task.

\subsection{Evaluation on nuScenes}

\subsubsection{Evaluation Metrics.} We adopt two widely used open-loop planning evaluation protocols on nuScenes: the ST-P3 protocol~\cite{hu2022st} and the UniAD protocol~\cite{hu2023planning}. Both protocols evaluate planned trajectories over 1s, 2s, and 3s future horizons using two core metrics: \textbf{L2 error}, which measures the Euclidean distance between predicted and ground-truth trajectory waypoints, and \textbf{collision rate}, which computes the frequency of collisions between the ego vehicle's occupied area along the planned trajectory and the bounding boxes of surrounding agents.

\subsubsection{Quantitative Analysis.} We further evaluate ExploreVLA on the nuScenes dataset to demonstrate the performance of our model. Note that we only apply Stage 1 (pre-training and supervised fine-tuning) without Stage 2 reinforcement learning post-training, as nuScenes lacks well-established closed-loop evaluation metrics that could serve as effective reward signals for RL optimization. As shown in \cref{tab:nuscenes_quantitative}, we compare ExploreVLA against state-of-the-art methods under both the ST-P3~\cite{hu2022st} and UniAD~\cite{hu2023planning} evaluation protocols. Under the ST-P3 protocol, ExploreVLA achieves competitive L2 errors (0.44m average) while attaining the lowest average collision rate of 0.10\%, matching OpenDriveVLA~\cite{zhou2025opendrivevla} and substantially outperforming other baselines. Notably, our model achieves the best collision rate at 1s and 2s horizons, indicating strong short-term safety-aware planning. Under the UniAD protocol, ExploreVLA also obtains reasonable L2 errors (0.77m average), remaining competitive with established methods such as UniAD (1.03m) and AutoVLA (0.70m). These results confirm that even without RL-based post-training, our Stage 1 model already learns robust driving representations that perform consistently well across benchmarks.

\begin{table}[tb]
  \caption{\textbf{Comparison on the nuScenes dataset.} The best performance is marked in \textbf{bold}.}
  \label{tab:nuscenes_quantitative}
  \centering
  \resizebox{\textwidth}{!}{
  \setlength{\tabcolsep}{3pt}
  \begin{tabular}{lcccccccccccccccc}
    \toprule
    \multirow{3}{*}{Model} 
    & \multicolumn{8}{c}{\textbf{ST-P3 metrics}~\cite{hu2022st}} 
    & \multicolumn{8}{c}{\textbf{UniAD metrics}~\cite{hu2023planning}} \\
    \cmidrule(lr){2-9} \cmidrule(lr){10-17}
    & \multicolumn{4}{c}{L2 (m) ↓} 
    & \multicolumn{4}{c}{Collision (\%) ↓} 
    & \multicolumn{4}{c}{L2 (m) ↓} 
    & \multicolumn{4}{c}{Collision (\%) ↓} \\
    \cmidrule(lr){2-5} \cmidrule(lr){6-9} \cmidrule(lr){10-13} \cmidrule(lr){14-17}
    & 1s & 2s & 3s & Avg. 
    & 1s & 2s & 3s & Avg. 
    & 1s & 2s & 3s & Avg. 
    & 1s & 2s & 3s & Avg.\\
    \midrule

    ST-P3~\cite{hu2022st} & 1.33 & 2.11 & 2.90 & 2.11 & 0.23 & 0.62 & 1.27 & 0.71 & -& -& -& -& -& -& -& -\\
    VAD~\cite{jiang2023vad}  & 0.17 & 0.34 & 0.60 & 0.37 & 0.07 & 0.10 & 0.24 & 0.14  & -& -& -& -& -& -& -& - \\
    UniAD~\cite{hu2023planning} & 0.44 & 0.67 & 0.96 & 0.69 & 0.04 & 0.08 & 0.23 & 0.12 & 0.48 & 0.96 & 1.65 & 1.03 & 0.05 & \textbf{0.17} & 0.71 & 0.31\\
    EMMA~\cite{hwang2024emma}& \textbf{0.14} & \textbf{0.29} & \textbf{0.54} & \textbf{0.32} & -& -& -& - & -& -& -& -& -& -& -& - \\
    OpenEMMA~\cite{xing2025openemma} & 1.45 & 3.21 & 3.76 & 2.81 & -& -& -& - & -& -& -& -& -& -& -& - \\
    OpenDriveVLA~\cite{zhou2025opendrivevla} & \textbf{0.14} & 0.30 & 0.55 & 0.33 & 0.02 & 0.07 & \textbf{0.22} & \textbf{0.10} & \textbf{0.19} & \textbf{0.58} & 1.24 & \textbf{0.67} & \textbf{0.02} & 0.18 & 0.70 & \textbf{0.30}\\
     AutoVLA~\cite{zhou2025autovla} & 0.21 & 0.38 & 0.60 & 0.40 & 0.13 & 0.18 & 0.28 & 0.20  & 0.28 & 0.66 & \textbf{1.16} & 0.70 & 0.14 & 0.25 & \textbf{0.53} & 0.31\\ 
    \midrule
     ExploreVLA & 0.28 & 0.40 & 0.65 & 0.44 & \textbf{0.01} & \textbf{0.05} & 0.25 & \textbf{0.10}  & 0.31 & 0.64 & 1.37 & 0.77 & - & - & - & -\\ 
    \bottomrule
  \end{tabular}
  }
\end{table}

\subsection{Closed-Loop Evaluation on HUGSIM}
\label{sec:hugsim}

While our main experiments adopt the non-reactive NAVSIM benchmark and the open-loop nuScenes benchmark, we additionally assess ExploreVLA under a fully reactive closed-loop setting to examine how it behaves when its own actions drive the evolution of the scene. We evaluate on HUGSIM~\cite{zhou2025hugsim}, a photo-realistic closed-loop simulator built on 3D Gaussian Splatting that spans
four source domains (KITTI-360, Waymo, nuScenes, and PandaSet). Performance is measured by the HD-Score~\cite{zhou2025hugsim}, which aggregates
no-collision, drivable-area compliance, time-to-collision, comfort, and route completion into a single value in $[0,1]$. We conduct \emph{zero-shot} evaluation, i.e., the model is trained as described in the main paper and directly deployed in HUGSIM without any closed-loop fine-tuning or domain adaptation.

\begin{table}[t]
  \centering
  \caption{Zero-shot closed-loop evaluation on HUGSIM, reported as HD-Score ($\uparrow$) on four source domains. The best result per domain is marked in \textbf{bold} and the second best is \underline{underlined}. ExploreVLA is trained as in the main paper and deployed without any closed-loop fine-tuning.}
  \label{tab:hugsim}
  \setlength{\tabcolsep}{3pt}
  \begin{tabular}{lcccc}
    \toprule
    Method & KITTI-360 $\uparrow$ & Waymo $\uparrow$ & nuScenes $\uparrow$ & PandaSet $\uparrow$ \\
    \midrule
    VAD~\cite{jiang2023vad}    & 0.028 & 0.154 & 0.368 & 0.428 \\
    UniAD~\cite{hu2023planning} & \underline{0.071} & \textbf{0.581} & 0.399 & 0.771 \\
    LTF~\cite{chitta2022transfuser}              & 0.054 & \underline{0.480} & \textbf{0.626} & \textbf{0.857} \\
    \midrule
    ExploreVLA        & \textbf{0.105} & 0.360 & \underline{0.517} & \underline{0.821}\\
    \bottomrule
  \end{tabular}
\end{table}

As shown in Tab.~\ref{tab:hugsim}, ExploreVLA performs competitively against strong closed-loop baselines despite never being optimized in a closed-loop setting. It achieves the best HD-Score on KITTI-360 and ranks second on both nuScenes and PandaSet. These results indicate that the dense world-modeling supervision and uncertainty-guided exploration learned during training transfer reasonably well to reactive closed-loop conditions.

We also observe a notable gap on the Waymo domain. We attribute this primarily to the open-loop nature of our RL fine-tuning: because the policy is refined from ground-truth states, it receives limited experience in recovering from the self-induced, rarely-seen states that arise under closed-loop rollout, and the exploration reward may still anchor the policy toward the training distribution. We regard closed-loop RL fine-tuning as a promising direction for future work.

\end{document}